\documentclass[11pt,a4paper]{article}
\usepackage{naaclhlt2019}
\usepackage{versions}\excludeversion{hide}
\usepackage[usestackEOL]{stackengine}
\renewcommand{\shortstack}[1]{\Centerstack{#1}}
\usepackage{footnote}
\usepackage{xspace}
\newcommand{\fastText}{\texttt{\small  fastText}\xspace}
\usepackage[T1]{fontenc}
\usepackage[utf8x]{inputenx}
\newcommand{\RL}[1]{\beginR #1\endR}
\newfont{\hebrew}{Jerusalem} 

\newcommand{\heb}[1]{{\RL{\hebrew #1}}}    
\newcommand{\word}[2]{``#2'' (\heb{#1})}
\usepackage{times}
\usepackage{latexsym}
\usepackage{graphicx}
\usepackage{tikz-dependency}
\aclfinalcopy 
\def\aclpaperid{***} 

\title{Semantic Characteristics of Schizophrenic Speech}

\makeatletter
\def\@fnsymbol#1{\ensuremath{\ifcase#1\or *\or \ast\ast \else\@ctrerr\fi}}
\makeatother

\author{Kfir Bar\thanks{$^\ast$Equal contribution.} \\
  School of Computer Science \\
   College of Management\\ Academic Studies \\
  Rishon LeZion, Israel \\
  {\tt kfirb@colman.ac.il} 
  \And
  Vered Zilberstein$^{\ast,\!\!}$ \thanks{$^\ast{^\ast}$Supported by the Deutsch Institute.}\\
  School of Computer Science \\
  Tel Aviv University \\
  Ramat Aviv, Israel \\
  {\tt veredz1@mail.tau.ac.il} 
  \And
  Ido Ziv$^\ast$\\
  Department of Psychology \\
   College of Management\\ Academic Studies \\
  Rishon LeZion, Israel \\
  {\tt idoz@colman.ac.il} 
  \AND
  Heli Baram$^\ast$ \\
  Department of Psychology \\
   Ruppin Academic Center\\
  Emek Hefer, Israel \\
  {\tt fanta.hchc@gmail.com}
  \And
  Nachum Dershowitz \\
  School of Computer Science \\
  Tel Aviv University\\
  Ramat Aviv, Israel \\
  {\tt nachum@tau.ac.il}
  \AND
  Samuel Itzikowitz \\
  School of Computer Science \\
  College of Management Academic Studies \\
  Rishon LeZion, Israel \\
  {\tt samitz@st.colman.ac.il}
  \And
  Eiran Vadim Harel \\
  Beer Yaakov  Mental Heath Center \\
  Beer Yaakov, Israel \\
  {\tt eiran.harel@moh.gov.il}
}
\date{\today}
\begin{document}
\maketitle
\begin{abstract}
Natural language processing tools are used to automatically detect disturbances in transcribed speech of schizophrenia inpatients who speak Hebrew.
We measure topic mutation over time and show that controls maintain  more cohesive speech than  inpatients. 
We also examine differences in how inpatients and controls use adjectives and adverbs to describe content words and show that the ones used by  controls are more common than the those of inpatients.
We provide  experimental results and show their potential for automatically detecting schizophrenia in  patients  by  means only of their speech patterns. 
\end{abstract}

\section{Introduction}
Thought disorders are described as disturbances in the normal way of thinking. 
Bleuler \shortcite{Bleuler91} original considered thought disorders to be a speech impairment in schizophrenia patients, but nowadays there is agreement that thought disorders are also relevant to other clinical disorders, including pediatric neurobehavioral disorders like attention deficit hyperactivity disorder 
and high functioning autism. 
They can even occur in normal populations, especially in people who have a high level of creativity.
Bleuler focused mostly on ``loosening of associations'', or \textit{derailment}, a thought disorder  characterized by the usage of unrelated concepts in a conversation, or in other words, a conversation  lacking coherence. 
The \textit{Diagnostic and Statistical Manual of Mental Disorders
(DSM 5)} \cite{DSM:5} outlines \textit{disorganized speech} as one of the criteria for making a diagnosis of schizophrenia. 
Morice and Ingram \shortcite{doi:10.3109/00048678209161186} showed that schizophrenics' speech is built upon a different syntactic structure than normal controls, and that this difference increases over time. 
Andreasen \shortcite{andreasen1979thought} suggested several definitions of linguistic and cognitive behaviors frequently observed in patients, and which may be useful for thought-disorder evaluation.
Among the definitions presented in that report, one  finds the following, which we  address in this study:

\noindent\textbf{Incoherence}, also known as ``word salad'',  refers to speech that is incomprehensible at times due to multiple grammatical and semantic inaccuracies. 
In this paper, we focus mostly on the semantic inaccuracies, leaving grammatical issues for future investigation.

\noindent\textbf{Derailment}, also known as ``loose associations'',  happens when a speaker shifts among topics that are only remotely related, or are completely unrelated, to the previous ones.

\noindent\textbf{Tangentiality} occurs when an irrelevant, or just barely relevant, answer is provided for a given question. 

We focus here on derailment.
But tangentiality has been addressed in some other studies. 
The two  notions are closely related.

One of the main data sources for diagnosing mental disorders is speech, typically collected during a psychiatric interview. 
Identifying signals that indicate the presence of thought disorders is often  challenging and subjective, especially in patients who are not undergoing a psychotic episode at the time of the interview.

In this work, we focus on schizophrenia.
We investigate a number of semantic characteristics of transcribed human speech, and propose a way to use them to measure disorganized speech. 
Natural-language processing  software is used to automatically detect those characteristics, and we suggest a way of aggregating them in a meaningful way. 
We use transcribed interviews, collected from Hebrew-speaking schizophrenia inpatients at a mental health hospital and from a control group. 
About two thirds of the patients were identified as in schizophrenia remission at the time of the interview.

Following a few previous works \citep{W18-0615, bedi2015automated}, we measure Andreasen’s derailment by calculating average semantic similarity between consecutive chunks of a running text to track topical mutations, and show the difference between  patients and  controls. 
For incoherence, we look at word modifiers, focusing on adjectives and adverbs, that  subjects use to describe the same objects, and then learn the difference between the two groups. 
As a final step, we use those semantic characteristics in a classification setting and argue for their usability.

This work makes the following contributions:
\begin{itemize}
\item We measure derailment in speech using word semantics, similar to \citep{bedi2015automated}, this time on Hebrew.
\item We explore a novel way of measuring one aspect of speech incoherence, by measuring how similar modifiers (adjectives and adverbs) are to ones used in a reference text to describe the same words.
\item
Using these measures, we build a classifier for detecting schizophrenia on the basis of recorded interviews, which  achieves 81.5\% accuracy.
\end{itemize}

We proceed as follows: The next section reviews some relevant previous work. In Section~\ref{sect:data_collection}, we describe how we collected the data. Our main contributions are described in Section~\ref{sect:tools_method}, followed by some conclusions suggested in the final section.

\section{Related Work}
There is a large body of work that examines human-generated texts with the aim of learning about the way people who suffer from various mental-health disorders use language in different settings. 
For example, Al-Mosaiwi and Johnstone \shortcite{mosaiwi}  conducted a study in which they analyzed 63 web forums, some  related to  mental health disorders and  others  used as control. 
They ran their analysis with the well-known Linguistic Inquiry and Word Count 
\cite{pennebaker2015development} tool to find absolutist words in free text. 
Overall, they discovered that anxiety, depression, and suicidal-ideation forums contained more absolutist words than  control forums. 

Recently, social media have become a vital source for learning about how people who suffer from mental-health disorders use language. 
Several studies collect relevant users from Twitter,\footnote{\url{https://twitter.com}} by considering users who intentionally write about their diagnosed mental-health disorders. 
For example, in  \citep{de2013predicting, tsugawa2015recognizing}, some language characteristics of Twitter users who claim to suffer from a clinical depression are studied. 
Similarly, users who suffer from post traumatic stress disorder 
are addressed in \citep{Coppersmith2014MeasuringPT}. Mitchell et al.~\shortcite{Mitchell2015QuantifyingTL} analyze tweets posted by schizophrenics, and Coppersmith et al.~\shortcite{coppersmith2016exploratory} investigate the language and emotions that are expressed by users who have previously attempted to commit suicide. 
Coppersmith et al.~\shortcite{coppersmith2015adhd} work with users who suffer from a broad range of mental-health conditions and explore language differences between  groups. 
Most of these works found a significant difference in the usage of some linguistic characteristics by the experience group when compared to a control group. 
Furthermore, different levels of these linguistic characteristics are used as features for training a classifier to detect mental-health disorders prior to the report date.

Reddit\footnote{\url{https://www.reddit.com}} has also been identified as a convenient source for collecting data for this goal. 
Losada and Crestani~\shortcite{losada2016test} outline a methodology for collecting posts and comments of Reddit and Twitter users who suffer from depression. 
Similarly, a large dataset of Reddit users with depression, manually verified (by lay annotators for an explicit claim of diagnosis), has been released for public use \cite{DBLP:journals/corr/abs-1709-01848}. 
In that work, the authors employ a deep neural network on the raw text for detecting clinically depressed people ahead of time, achieving 65\% F1 score on an evaluation set.

A few caveats are in order when using social media for analyzing mental health conditions. 
First, self reporting of a mental health disorder is not a popular  course of action. 
Clearly, then, the experimental group is chosen from a subgroup of the relevant population. 
Second, the controls, typically collected randomly ``from the wild'', are not guaranteed to be free of mental-health disorders. 
Finally, social media posts are considered to be a different form of communication than ordinary speech.
For all these reasons, in this work, we use  validated experimental and control groups in an interview setting.

Measuring various aspects of incoherence in schizophrenics using computational tools has been previously addressed in \citep{elvevaag2007quantifying, bedi2015automated, W18-0615}. 
Elvevåg et al.~\shortcite{elvevaag2007quantifying} analyzed transcribed interviews of inpatients with schizophrenia to measure tangentiality. 
Moving along the patient's response, they calculated the semantic similarity between text chunks of different sizes and the question that was asked by the interviewer. 
Semantic similarity was cast by cosine similarity over the latent semantic  analysis (LSA) \citep{deerwester1990indexing} vectors calculated for each word, and summed across an entire chunk of words.
They fitted a linear-regression line to represent the trend of the cosine similarity values, as one moves along the text. 
The slope of that line was used to measure how quickly the topic diverges from the original question. 
Overall, they were able to show a significant correlation between those values and a blind human evaluation of the same responses. 
Furthermore, as  chunk size grows larger, the distinction between patients and controls becomes less prominent. 
One explanation for that could be the large number of mentions of functional and filler words, for which we typically do not have a good semantic representation. 
Iter et al.~\shortcite{W18-0615} addressed this suggestion by cleaning the patients' responses of all those words and expressions (e.g.\@ \textit{uh}, \textit{um}, \textit{you know}) prior to calculating the semantic scores.
This gave a slight improvement, although measured over a relatively small set of participants. 
Instead of working with  chunks of text, they worked with full sentences, and replaced LSA with some modern techniques for sentence embeddings. 
Likewise, in our work, we use word embeddings instead of LSA.

Bedi et al.~\shortcite{bedi2015automated} define coherence as an aggregation of the cosine similarity between pairs of consecutive sentences, each represented by the element-wise average vector of the  individual words' LSA vectors. 
They worked with a group of 34 youths at clinical high-risk for psychosis, interviewed them quarterly for 2 1/2 years, and transcribed their answers. 
Five out of the 34 transitioned to psychosis. 
They used  coherence scores, along with part-of-speech information, to automatically predict transition to psychosis with 100\% accuracy. 

The goal of all these works, including ours, is to automatically detect disorganized speech in a more objective and reliable way. 
Inspired by the last three studies described above, we analyzed transcribed responses to 18 open questions given by inpatients with schizophrenia and by controls. 
Instead of cleaning the text from filler words using a dictionary -- as proposed by \cite{W18-0615}, we take a deeper look into the syntactic roles the words play, and calculate semantic similarity over a filtered version of the text, every time using different sets of part-of-speech categories. 
We report on the results of two sets of experiments:
(1) We measure derailment by calculating the semantic similarity of adjacent words of various part-of-speech categories. 
(2) We measure semantic coherence by looking at the choices of modifiers (adjectives, adverbs) used in responses by inpatients and controls, as compared to those used in ordinary discourse.   

Generally speaking, not too much is known about the role played by adjectives and adverbs in thought disorders. 
Modifiers are often not included in language tests, as they usually need to be presented together with the  noun or verb they modify. 
Some previous works \citep{obrkebska2007lexical} have reported a significantly smaller number of adjectives used by schizophrenics. 
In the current study, we use computational tools to investigate the semantic relation between modifiers and objects, and its attribution to speech incoherence. 

\section{Data Collection}
\label{sect:data_collection}
We interviewed 51 men, aged 19--63, divided into control and patient groups, all speaking Hebrew as their mother tongue. 
The patient group comprised 24 inpatients at Beer Yaakov Mental Health Center in Israel who were officially diagnosed with schizophrenia. 
The control group includes 27 people, mainly recruited via an advertisement that we placed on social media. 
Most of the participants are single, with average-to-lower monthly income. 
Demographics for the two groups are presented in Table~\ref{tab:demographics}.

\begin{table}[t!]
\begin{center}
\begin{tabular}{|l|c|c|}
\hline 
\bf  & \bf Control & \bf Patients \\ \hline
\hline
$N$ & 27 & 24\\
\hline
Age, Mean (SD) & 30.3 (8.26) & 38.3 (10.43)\\
\hline
Edu., HS & 68\% & 75\% \\
\hline
Edu., Post HS & 20\% & ~4\% \\
\hline
Loc., South & 40\% & 20\% \\
\hline
Loc., Center & 44\% & 33\% \\
\hline
M.S., Single & 80\% & 95\% \\
\hline
Income, Avg/low & 84\% & 83\% \\
\hline
\end{tabular}
\end{center}
\caption{\label{tab:demographics} Demographics by group. Edu. = Education (HS = High School); Loc. = Location in Israel; M.S. = Marital Status.}
\end{table}

\par\noindent{\small\textbf{Ethics statement:} The institutional review board of the College of Management Academic Studies of Rishon LeZion, as well as of the Beer Yaakov--Ness Ziona Mental Health Center, approved these experiments, and informed consent was obtained for all subjects.}

\subsection{Interviews}

Overall, the participants were asked 18 questions, out of which 14 were thematic-apperception-test (TAT) pictures that  participants were requested to describe, followed by 4 questions that require the participant to share some personal thoughts and emotions. 
Both the control and patient groups completed a demographic questionnaire. 
To monitor the mental-health condition of the control group, they were requested to complete Beck's Depression Inventory-II (BDI-II) and the State and Trait Anxiety Inventory (STAI).
The patient group  also completed BDI-II, as well as a Hebrew translation \cite{KATZ2012850} of the Positive and Negative Syndrome Scale--6 (PANSS-6, a shorter version of PANSS-30) questionnaire, in order to assess symptoms of psychosis \cite{doi:10.1111/acps.12526}.
Scores for the two questionnaires were found to be highly correlated.
Out of the patient group, 66.7\% were assigned a score below 14, a recommended preliminary threshold indicating schizophrenia remission.

\begin{figure}
    \centering
    \includegraphics[width=8cm,trim=0 4mm 0 13mm,clip]{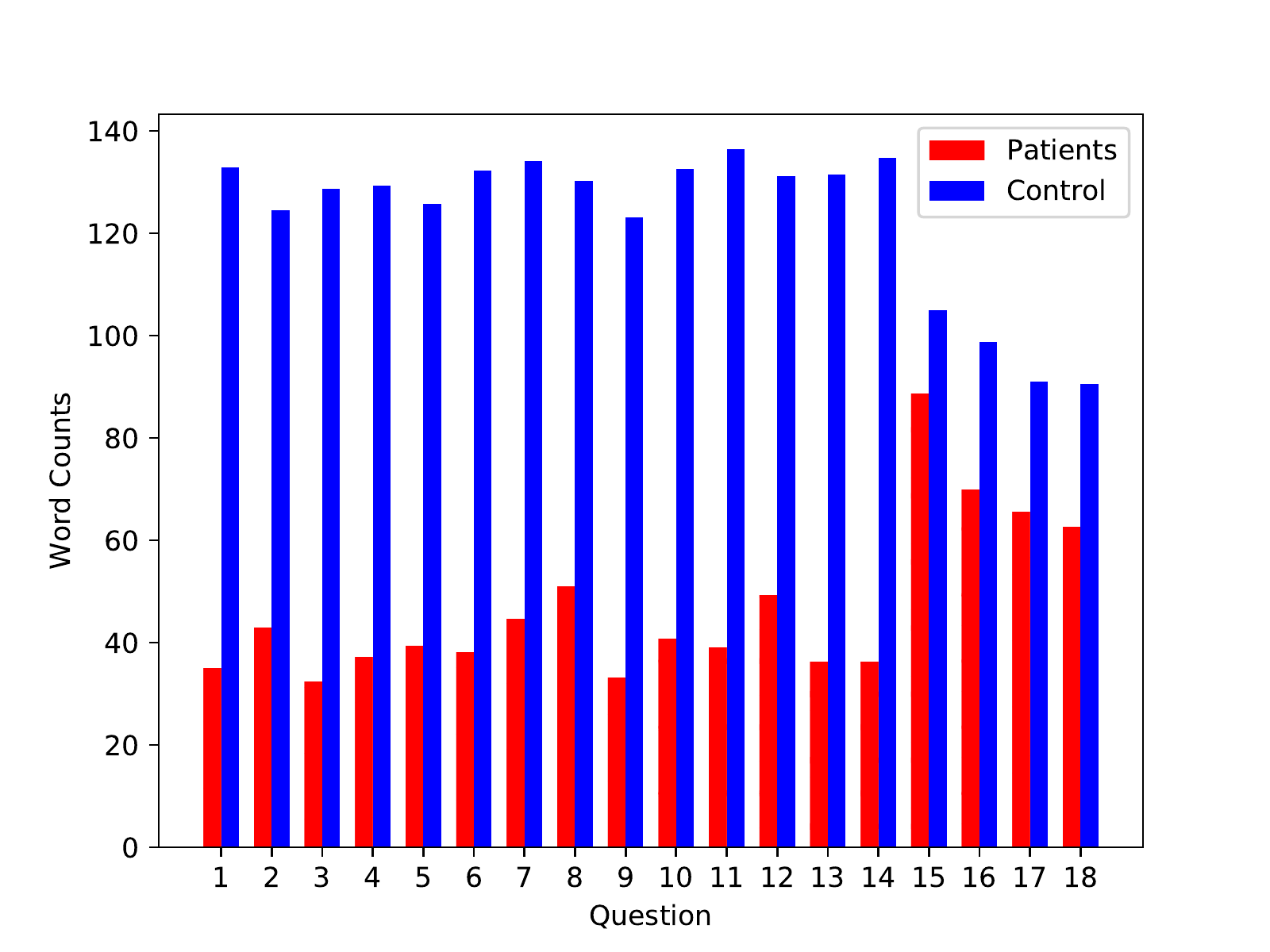}
    \caption{Word counts per question.}
    \label{fig:word_count}
\end{figure}

The interviews were recorded and then manually transcribed by Hebrew-speaking students from our lab. 
The TAT pictures  presented to participants during the interview were: 1, 2, 3BM, 4, 5, 6BM, 7GF, 8BM, 9BM, 12M, 13MF, 13B, 14, 3GF. 
Table~\ref{tab:questions} lists the questions that were presented to the participants during the interview.
All the transcripts are written in Hebrew. 
Figure~\ref{fig:word_count} shows average word counts by question, per group. 
Clearly, the patients spoke fewer words than the controls. 
The difference becomes less significant for the open-ended questions.

\begin{table*}[t!]
\begin{center}
\begin{tabular}{|c|l|c|}
\hline 
\bf ID & \bf Question \\ \hline
\hline
1 & Tell me as much as you can about your bar mitzvah. \\
\hline
2 & What do you like to do, mostly?\\
\hline
3 & What are the things that annoy you the most?\\
\hline
4 & What would you like to do in the future?\\
\hline
\end{tabular}
\end{center}
\caption{\label{tab:questions} Four open questions asked during the interview.}
\end{table*}

\subsection{Preprocessing}

Hebrew being a highly-inflected language, we preprocessed the texts with the Ben-Gurion University Morphological Tagger \cite{bgut07}, a context-sensitive morphological analyzer for Modern Hebrew. 
Given a running text, the tagger breaks the text into words and provides morphological information for every word, including the disambiguated part-of-speech tag and lemma. 
There were no specific instructions given to the transcribers for how to punctuate,
which led  to an inconsistency in the way punctuation  was used in the transcriptions. 
We used the tags to clean up all punctuation marks by removing all tokens tagged as such.

\section{Tools and Method}
\label{sect:tools_method}
We report on two sets of experiments.
In the first, we measure derailment by calculating the semantic similarity between adjacent words in running text. 
In the second set of experiments, we investigate the modifiers that the two groups  use to describe specific nouns and verbs.
As a final step, we measure the contribution of the semantic characteristics that we compute in the experiments, for automatic classification of schizophrenia.

\subsection{Experiment 1: Measuring Derailment}
\label{sect:exp1}
We calculate a derailment score for each response and use it to measure derailment.

\noindent\textbf{Tools:} To measure derailment, we calculate the semantic similarity of adjacent words in the answers provided by the participants during the interview. 
We use word embeddings to represent each word by means of a mathematical vector that captures its meaning.
These vectors were created automatically by characterizing words by the surrounding contexts in which they are mentioned in a large corpus of documents.
Specifically, we used Hebrew pretrained vectors provided by \fastText \cite{grave2018learning}, which were created from Wikipedia,\footnote{\url{https://www.wikipedia.org}} as well as from other content  extracted from the web with Common Crawl.\footnote{\url{http://commoncrawl.org}} 
Overall, 97\% of the words in our corpus exist in \fastText. 
Hebrew words are inflected for person, number and gender; prefixes and suffixes are added to indicate definiteness, conjunction, prepositions, and possessive forms. 
On the other hand, \fastText was trained for surface forms.
Therefore, we work on the surface-form level.
To measure semantic similarity between two words, we use the common cosine-similarity function that calculates the cosine of the angle between the two corresponding vectors. 
The score ranges from $-1$ to $+1$, with $+1$ representing maximal similarity.

\noindent\textbf{Method:} (1) For each sufficiently long response, $R$, we retrieve the \fastText vector $v_i$ for every word $R_i$, $i=0\dots n$, in the response. 
(2) For each word, we calculate the average pairwise cosine similarity between this word and the $k$ following words. The integer $k$ is a parameter; we experimented with different values.
(3) We take the average of all the individual cosine similarity scores and form a single score for each response.

In this experiment, we consider only responses that are long enough to allow topic mutation to develop. 
Therefore, we use only the four questions from Table~\ref{tab:questions} for which the participants provided a relatively long response. 
Accordingly, we drop responses of fewer than 50 words.
As mentioned above, we consider that the existence of some word types, like fillers and functional words, might introduce some noise, which might harm the calculation process. 
We would rather focus on words that convey real content.
Therefore, we calculate  scores separately using all words and  using only  \textit{content words},
which we take to be nouns, verbs, adjectives, and adverbs.
We detected a few types of text repetitions, which may bias the derailment score. 
One type is when a word is said twice or more for emphasis; for example, 
\word{מהר מהר}{quickly, quickly} 
(i.e.\@ very quickly). 
To mitigate this bias, we  keep only one word out of a pair of consecutive identical words.
Another type is when a whole phrase is repeated; for example,
\word{היא ממהרת מאוד, היא ממהרת מאוד}%
{She's in a big hurry; she's in a big hurry}.
Handling this problem is left for future work.

We calculate derailment scores for the responses provided by all participants and compare the means of the two groups.

\begin{figure*}
    \centering
    \includegraphics[width=12.7cm]{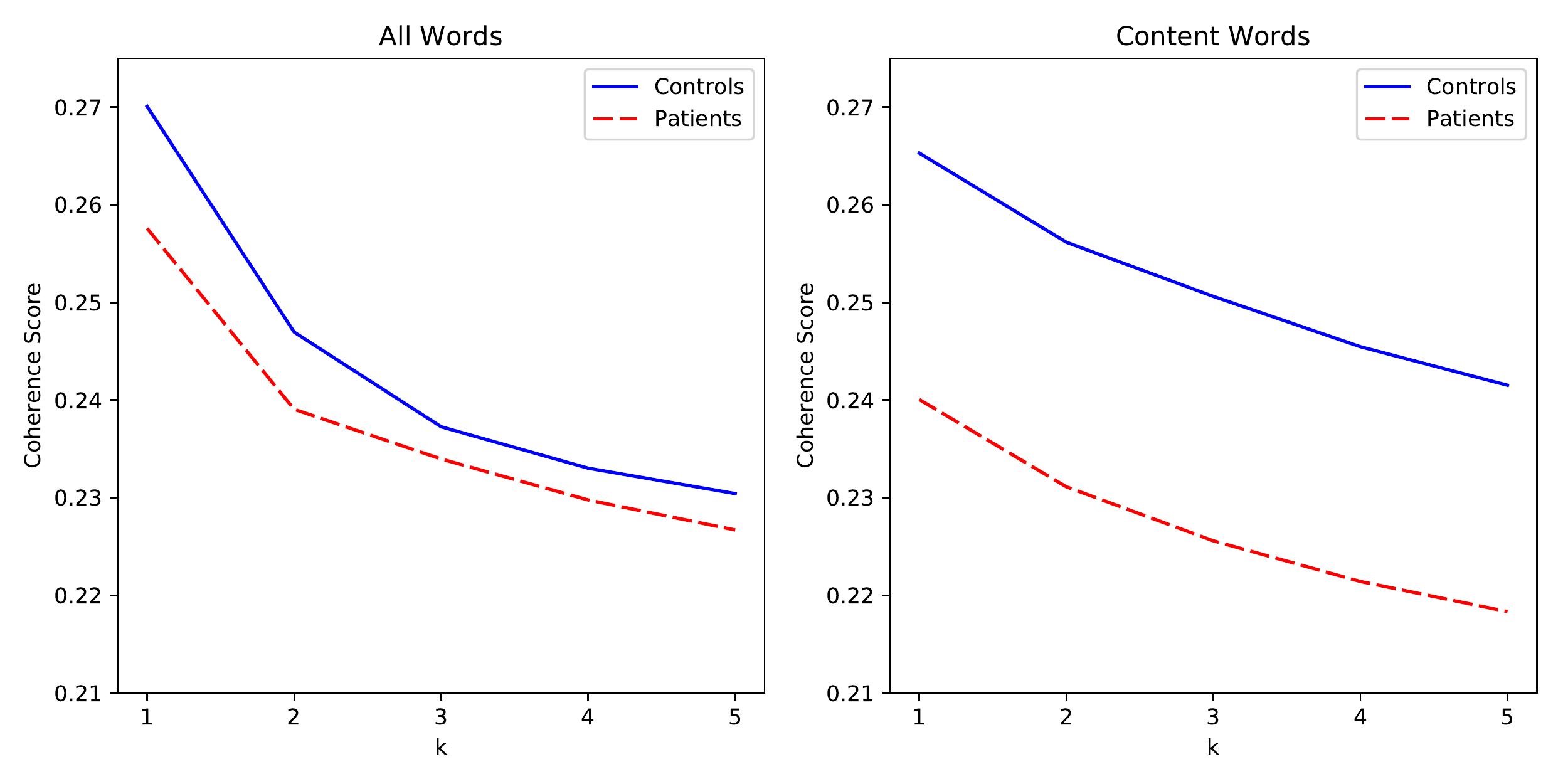}
    \caption{Derailment scores for different values of $k$. The left plot shows the results for all word types, and the right plot shows the results for content words only.}
    \label{fig:coherence}
\end{figure*}

\noindent\textbf{Results:}
When using all words, we could not detect a significant difference between  patients and controls. 
However, when using content words only, patients scored lower on derailment than the controls, for all window widths $k$, suggesting that focusing only on content words is the more robust approach for calculating derailment. 
This finding is consistent with previous work \citep{W18-0615}.
Overall,  coherence  decreases as $k$ increases. 
Table~\ref{tab:exp1_res} summarizes the results. To confirm the significance we are seeing in the results, is due to the diagnosis and not due to other characteristics of the participants, we aggregated the same scores for the different age groups and education levels, regardless of the diagnosis status; all these results did not appear to be significant.
Figure~\ref{fig:coherence} shows the trend of the average derailment score from Table~\ref{tab:exp1_res}, running with different values of $k$.  
The left plot was produced for all word types, and the right plot using only content words. 
We clearly observe a slight increase of the entire control curve and a slight decrease of the patients curve, when restricting to content words.

\begin{table*}[t!]
\begin{center}
\begin{tabular}{|c|ccc|ccc|}
\hline
\multicolumn{1}{|c|}{} & \multicolumn{3}{c|}{\textbf{All words}} & \multicolumn{3}{c|}{\textbf{Content words}} \\ \hline \hline
$k$ & Control & Patients & $t$ & Control & Patients & $t$ \\ \hline
1 & \shortstack{0.270 \\ (0.014)} & \shortstack{0.257 \\ (0.025)} & 2.004* & \shortstack{0.265 \\ (0.019)} & \shortstack{0.240 \\ (0.020)} & 2.968* \\\hline
2 & \shortstack{0.246 \\ (0.017)} & \shortstack{0.239 \\ (0.025)} & 1.173 & \shortstack{0.256 \\ (0.018)} & \shortstack{0.231 \\ (0.025)} & 2.687* \\\hline
3 & \shortstack{0.237 \\ (0.017)} & \shortstack{0.233 \\ (0.025)} & 0.476 & \shortstack{0.250 \\ (0.018)} & \shortstack{0.225 \\ (0.026)} & 2.614* \\\hline
4 & \shortstack{0.233 \\ (0.018)} & \shortstack{0.229 \\ (0.025)} & 0.471 & \shortstack{0.245 \\ (0.018)} & \shortstack{0.221 \\ (0.026)} & 2.539* \\\hline
5 & \shortstack{0.230 \\ (0.017)} & \shortstack{0.226 \\ (0.026)} & 0.528 & \shortstack{0.241 \\ (0.018)} & \shortstack{0.218 \\ (0.023)} & 2.598*\\\hline
\end{tabular}
\end{center}
\caption{\label{tab:exp1_res} Results for Experiment 1. Comparing average derailment scores of patients and controls. The numbers are provided as average across patients and controls, with standard deviation in parentheses, $*p<0.05$.}
\end{table*}

\subsection{Experiment 2: Incoherence}
\label{sect:syntax}
In this experiment, we examine the way patients use adjectives and adverbs (hereafter,  \textit{modifiers}) to describe specific nouns and verbs, respectively. 
Our goal is to measure the difference between  modifiers  used by  patients and the ones used by controls, when describing the same nouns and verbs.
We suggest this as a tool for measuring incoherence in speech.
For example, inspecting the responses for the first TAT image, we learn that patients typically use the adjectives
\word{חדש}{new}
and 
\word{טוב}{good}
to modify the noun 
\word{כינור}{violin}, while controls use the adjectives \word{ישן}{old}, \word{עצוב}{sad}, and \word{משמעותי}{significant}.

\begin{figure}
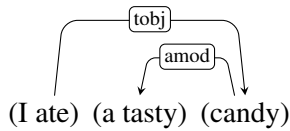

    \centering
    \begin{dependency}[theme=default]
        \begin{deptext}
         \&  (I  ate) \& (a tasty) \& (candy) \\
        \end{deptext}
        
        \depedge[->]{2}{4}{tobj}
        \depedge[->]{4}{3}{amod}
        \end{dependency}
    \caption{The dependencies returned by YAP for the sentence ``(I ate) (a tasty) candy''. The parentheses delimit the translations for each of the three Hebrew words in the sentence.}
    \label{fig:dep_example}
\end{figure}

\noindent\textbf{Tools}: To detect all noun-adjective and verb-adverb pairs in the responses, we use a {dependency parser}, which analyzes the grammatical structure of a sentence and builds links between ``head'' words and their modifiers. 
Specifically, we use YAP \cite{moretsarfatycoling2016}, a dependency parser for Modern Hebrew, and process each sentence individually. 
Among other things, YAP provides a word-dependency list, shaped as a list of tuples, each includes a head word, a dependent word, and  the kind of dependency.
We use the relevant types (e.g.\@ \textit{advmod}, \textit{amod}) for finding all noun-adjective and verb-adverb pairs.
For example, Figure~\ref{fig:dep_example} shows the dependencies returned by YAP for the input sentence: \word{אכלתי סוכריה טעימה}{I ate a tasty candy}. 
From this sentence we extract the noun \word{סוכריה}{candy}, which is modified by the adjective \word{טעימה}{tasty}.

\begin{savenotes}
\begin{table*}[t!]
\begin{center}
\begin{tabular}{|l|p{7cm}|r|r|}
\hline 
\bf Corpus & \bf Description& \bf \# Documents& \bf \# Words \\ \hline
\hline
Doctors\footnote{\url{https://www.doctors.co.il}} & Articles from the Doctors medical website & 239 & 187,938\\
\hline
Infomed\footnote{\url{https://www.infomed.co.il}} & Question-and-answer discussions from the Infomed website's medical forum, January 2006 -- September 2007 & 749 & 128,090\\
\hline
To Be Healthy\footnote{\url{https://tobehealthy.co.il}} & Articles and forum discussions from the To Be Healthy (L'Hiyot Bari, 2b-bari) medical website & 137 & 112,839\\
\hline
HaAretz\footnote{\url{https://www.haaretz.co.il}} & News and articles from the HaAretz news website, 1991 & 4,920 & 250,399\\
\hline
\end{tabular}
\end{center}
\caption{\label{tab:external_corpora} The external Hebrew corpora used to collect modifiers of nouns and verbs that are typically used.}
\end{table*}

\noindent\textbf{Method:} To measure the difference between the modifiers that are used by patients and controls, we compare them to the modifiers that are commonly used to describe the same nouns and verbs. 
For example, given an answer with only one noun \word{כינור}{violin} that is modified by the adjective \word{עצוב}{sad}, we calculate a score that reflects how similar the adjective ``sad'' is to  adjectives that are typically used to describe a violin.

We take the following steps: 

\noindent
(1) We convert each sentence into a list of noun-adjective and verb-adverb pairs using YAP.

\noindent
(2) To compare each modifier with the modifiers that are typically used to describe the same noun or verb, we use  external corpora as reference. 
These were taken from various sources reflecting the health domain we are working in.%
\footnote{All were downloaded from MILA Knowledge Center for Processing Hebrew: \url{http://mila.cs.technion.ac.il/resources_corpora.html}.}  
Table~\ref{tab:external_corpora} lists the sources and the corresponding number of documents and words that they contain.
Each document in these sources was processed in exactly the same way to find all noun-adjective and verb-adverb pairs. 

\noindent
(3) Given a list of noun-adjective and verb-adverb pairs of one response, we calculate the similarity score of every modifier that describes a specific noun or verb with the set of modifiers describing exactly the same noun or verb in the reference corpus.
Looking at our example  above, we would want to calculate a similarity score between the adjective \word{ישן}{old} and all the adjectives that are used to describe \word{כינור}{violin} in the reference corpus.
Searching for instances of the same Hebrew word is challenging due to Hebrew's rich morphology. 
Hebrew words are inflected for person, number, and gender; prefixes and suffixes are added to indicate definiteness, conjunction, various prepositions, and possessive forms. 
Therefore, we work on the lemma (base-form) level. 
Most vowels in Hebrew are  not indicated in standard writing; therefore, Hebrew words tend to be ambiguous, and determining the correct lemma for a word is nontrivial. 
We use the lemmas provided by YAP.

Another challenge is how to compare a single modifier with a group of modifiers that were taken from the reference corpus.
We take the \fastText vectors of the modifiers that were extracted from the reference corpus and aggregate them into a single vector. 
Then, we take cosine similarity between the modifier from the response and the aggregated vector of the modifiers from the reference corpus. 
As an aggregation function, we use element-wise weighted average of the individual modifiers' \fastText vectors, and define the weights to be the inverse-document-frequency (IDF) score to account more for modifiers that describe the noun or verb more uniquely. 
We calculate IDF scores using the reference corpora.
For this purpose, a ``{qualified}'' word is a noun or verb that has an IDF score and that has at least one modifier linked to it in either the control or patient corpus. 
Most of the nouns and verbs are non-qualified; we only consider qualified words in this investigation.

\noindent
(4) For each response, we calculate two scores, individually. 
The adjective-similarity score is the IDF-weighted average of the individual adjective scores we calculate in the previous step.
Similarly, the adverb-similarity score is the IDF-weighted average of the individual adverb scores we calculate in the previous step.

\noindent
(5) To calculate a score on the participant level, we average the scores of all the individual responses provided by the participant.
\end{savenotes}


The output of this process is a pair of scores, one for adjectives and one for adverbs, calculated for each participant. 
The higher a score is, the more similar the modifiers are to ones that are typically used to describe the same noun or verb.

\noindent\textbf{Results:} Table~\ref{tab:incoherence_res} summarizes the results. Overall, controls have significantly higher scores for both modifier types, indicating a higher agreement on modifiers by the controls and external writers.  

\begin{table}[t!]
\begin{center}
\begin{tabular}{|l|c|c|c|}
\hline 
\bf  & \multicolumn{1}{|l|}{\bf Control} & \multicolumn{1}{|l|}{\bf Patients} & \bf $t$\\
\hline \hline 
\bf Adj & \shortstack{0.5891 \\(0.0301)} & \shortstack{0.5498 \\ (0.0284)} & 4.7765***\\
\hline
\bf Adv & \shortstack{0.6880 \\ (0.0251)}& \shortstack{0.6254 \\ (0.0709)} & 4.2961***\\
\hline
\end{tabular}
\end{center}
\caption{\label{tab:incoherence_res} Results for Experiment 2. The numbers are average coherence scores across patients and controls (with standard deviations); ***$p<0.001$. }
\end{table}

There are more nouns and adjectives than verbs and adverbs, as summarized in Table~\ref{tab:incoherence_counts}. 
On average, participants use more adjectives to describe nouns than adverbs to describe verbs. 
Controls use about 0.61 adjectives per noun, while patients use 0.84 adjectives on average. 
Similarly, patients use more adverbs to describe a verb on average than controls do. 
While patients use about 0.42 adverbs per verb, controls use only 0.23. 
However, these differences are not significant.

\begin{table}[t!]
\begin{center}
\begin{tabular}{|l|rr||rr|}
\hline 
 & \multicolumn{2}{|c||}{\bf Control} & \multicolumn{2}{|c|}{\bf Patients} \\
&  Total & Qual. & Total & Qual.\\
\hline \hline 
\bf Nouns & 934 & 226 & 242 & 90\\
\bf Adjectives & 573 & 371 & 204 & 127\\
\hline
\hline
\bf Verbs & 699 & 60 & 204 & 34\\
\bf Adverbs & 166 & 104 & 86 & 50\\
\hline
\end{tabular}
\end{center}
\caption{\label{tab:incoherence_counts} Experiment 2: Counts of nouns, verbs, and their modifiers, across the two groups. Qual. = Qualified.}
\end{table}

\subsection{Classification}
As a final step, we train several classifiers to distinguish between controls and patients. We represent participants with the characteristics we compute in the two experiments. 
Specifically, each subject is represented by the following: (1) noun and verb derailment scores; (2) coherence scores for 5 windows, using all words; and (3) coherence scores for 5 windows, using only content words. 
In total, we use 12 scores per subject. Each classifier was trained using a 10-fold cross-validation evaluation of prediction quality over the 51 participants.
For each classifier, we report on the overall prediction accuracy, as well as precision and recall for the prediction of the patients group. 
The classification algorithms we tried are Random Forest \citep{Breiman:2001:RF:570181.570182} and XGBoost \citep{Chen:2016:XST:2939672.2939785}, both  based on decision trees,
and, in addition, linear support vector machines (SVM) \citep{Cortes:1995:SN:218919.218929}. 
Table~\ref{tab:classification_results} summarizes the results per classifier with respect to the different metrics.

\begin{table}[t!]
\begin{center}
\begin{tabular}{|l|c|c|c|}
\hline 
\bf Classifier & \multicolumn{1}{|l|}{\bf Acc.} & \multicolumn{1}{|l|}{\bf Prec.} & \multicolumn{1}{|l|}{\bf Recall}\\
\hline \hline 
\bf Random Forest & \shortstack{81.5\%} & \shortstack{91.3\%} & \shortstack{71.8\%} \\
\hline
\bf XGBoost & \shortstack{80.5\%} & \shortstack{86.8\%} & \shortstack{73.1\%}\\
\hline
\bf SVM & \shortstack{70.4\%} & \shortstack{72.1\%} & \shortstack{47.3\%}\\
\hline
\end{tabular}
\end{center}
\caption{\label{tab:classification_results} Classification results for each classifier.}
\end{table}

We used the decision-tree based classifiers to calculate the most important features, that is, the ones that have the greatest impact on prediction decisions.
The most important features were found to be the two derailment scores, as expected.

\section{Conclusions}
\label{sect:discussion}
With  the aim of detecting speech disturbances,
we have analyzed transcribed Hebrew speech, produced by schizophrenia inpatients and compared it with those of controls.   
We believe that  speech produced during a psychiatric interview is a more reliable data source for detecting disturbances than are social media posts.

Generally speaking,  we find that patients talk significantly less in interviews than  controls do. 

In one experiment, we use word embeddings to detect derailment, that is, when a speaker shifts to a topic that is not strongly related to  previously discussed ones. 
The results  show that  controls have higher scores, indicating that they keep the topic more cohesive than patients do. 
These results are in line with previous studies on English \citep{bedi2015automated}, which showed that schizophrenics have a lower score, calculated by a similar mathematical procedure. 

In a second experiment, we examine the difference in how patients and controls use adjectives and adverbs to describe nouns and verbs, respectively. 
Our results show that the adjectives and adverbs that are used by the controls are more similar to the typical ones used to describe the same nouns and verbs. 
For now, we consider this difference as related to speech incoherence; however, we plan to continue investigating this direction in the near future, when more data become available.

Analyzing Hebrew is more challenging than analyzing English due to Hebrew's rich morphology, as well as the absence of written vowels. 
In the first experiment, we work with \fastText, which provides word embeddings on the surface-form level.
In the second, we used lemmata rather than the word surface forms, so we can find multiple instances of the same lexeme.

As we did not measure the IQ of participants, some of the results might, to a certain extent, be attributable to differences in intellect. 
Moreover, as can be seen in Table~\ref{tab:demographics}, about 20\% of the control participants have some sort of post high-school education, while most of the inpatients did not continue beyond high-school. 
We plan to address these questions in followup work. 
Another limitation that we are aware of is related to the classification results, as
the number of participants we use for training the classifiers might be considered  relatively small.

Overall, we found the semantic characteristics that we compute in this study to be beneficial for the task of detecting thought disorders in Hebrew speech.
We plan to collect speech samples from more subjects, and to continue to explore additional semantic -- as well as grammatical --  textual characteristics to support the automatic detection of various mental disorders.



\bibliography{naaclhlt2019}

\end{document}